\documentclass{article}

\usepackage[preprint]{corl_2025} % Uncomment for pre-prints (e.g., arxiv); This is like ``final'', but will remove the CORL footnote.
\usepackage{amssymb}
\usepackage{times}
\usepackage{multicol}
\usepackage{wrapfig}
\usepackage{xcolor}
\definecolor{brightgreen}{HTML}{00FF00}
\definecolor{brightred}{HTML}{ff0000}
\definecolor{springgreen}{HTML}{5aa236}

\definecolor{blue_1}{HTML}{6d9eeb}
\definecolor{blue_2}{HTML}{1155cc}
\definecolor{blue_3}{HTML}{0b5394}

\definecolor{red_1}{HTML}{ea9999}
\definecolor{red_2}{HTML}{e06666}
\definecolor{red_3}{HTML}{cc0000}

\definecolor{active_camera_color}{HTML}{fe0439}
\definecolor{wrist_camera_color}{HTML}{08abff}

\definecolor{banana_color}{HTML}{fcd000}

\usepackage{wrapfig}
\usepackage{booktabs}  % for better-looking tables
\BeforeBeginEnvironment{wrapfigure}{\setlength{\intextsep}{0pt}}
\AfterEndEnvironment{wrapfigure}{\setlength{\intextsep}{0pt}}

\usepackage{graphics} % for pdf, bitmapped graphics files
\usepackage{epsfig} % for postscript graphics files
\usepackage{mathptmx} % assumes new font selection scheme installed
\usepackage{amsmath} % assumes amsmath package installed
\usepackage{amssymb}  % assumes amsmath package installed
\usepackage{siunitx} % for SI units
\usepackage{textcomp}
\usepackage{gensymb} % for degree command
\usepackage{booktabs}
\usepackage{multirow}
\usepackage{xcolor}
\usepackage{algorithm}
\usepackage{algpseudocode}

\usepackage{enumitem}
\usepackage{xspace}
\usepackage{float}
\usepackage{comment}
\usepackage{mathtools}
\usepackage{soul}
\usepackage{natbib}
\usepackage{wrapfig}

\usepackage{makecell}
\usepackage{rotating}
\usepackage[percent]{overpic}
\usepackage{contour}
\usepackage{courier}

\usepackage{subcaption} % issues a warning with CVPR/ICCV format
\usepackage[font=small]{caption}
\usepackage[percent]{overpic}

\usepackage[11pt]{moresize}

\usepackage{pifont}% http://ctan.org/pkg/pifont
\definecolor{MyDarkBlue}{rgb}{0,0.08,1}
\definecolor{airforceblue}{rgb}{0.36, 0.54, 0.66}
\definecolor{MyDarkGreen}{rgb}{0.02,0.6,0.02}
\definecolor{MyDarkRed}{rgb}{0.8,0.02,0.02}
\definecolor{MyDarkOrange}{rgb}{0.40,0.2,0.02}
\definecolor{MyPurple}{RGB}{111,0,255}
\definecolor{MyRed}{rgb}{1.0,0.0,0.0}
\definecolor{MyGold}{rgb}{0.75,0.6,0.12}
\definecolor{MyDarkgray}{rgb}{0.66, 0.66, 0.66}
\definecolor{MyPink}{rgb}{0.9, 0.33, 0.5}
\definecolor{MyCyan}{rgb}{0., 0.4, 0.4}

\definecolor{guidance_green}{RGB}{12,131,27}
\definecolor{theme_orange}{RGB}{255,138,0}
\definecolor{theme_blue}{RGB}{67,99,216}
\definecolor{theme_taro}{RGB}{219,176,234}

\definecolor{pure_green}{RGB}{0,255,0}
\definecolor{pure_red}{RGB}{255,0,0}

\makeatletter
\DeclareRobustCommand\onedot{\futurelet\@let@token\@onedot}
\def\@onedot{\ifx\@let@token.\else.\null\fi\xspace}

\def\eg{\emph{e.g}\onedot} 
\def\ie{\emph{i.e}\onedot}

\def\wrt{w.r.t\onedot} 

\makeatother

\newcommand{\oursfull}{Vision in Action}
\newcommand{\ours}{ViA}

\usepackage[most]{tcolorbox}

\usepackage{xparse}
\usepackage{lipsum}

\def\exampletext{Example} % If English

\NewDocumentEnvironment{testexample}{ O{} }
{
    \colorlet{colexam}{theme_blue} % Global example color
    \newtcolorbox[use counter=testexample]{testexamplebox}{%
        empty,% Empty previously set parameters
        title={\exampletext: #1},% use \thetcbcounter to access the testexample counter text
        attach boxed title to top left,
        minipage boxed title,
        boxed title style={empty,size=minimal,toprule=0pt,top=2pt,left=2mm,overlay={}},
        coltitle=colexam,fonttitle=\bfseries,
        before=\par\nonskip\noindent,parbox=false,boxsep=0pt,left=2mm,right=0mm,top=2pt,breakable,pad at break=0mm,
        before upper=\csname @totalleftmargin\endcsname0pt, % Use instead of parbox=true. This ensures parskip is inherited by box.
        after=\par\nonskip,
        overlay unbroken={\draw[colexam,line width=.5pt] ([xshift=-0pt]title.north west) -- ([xshift=-0pt]frame.south west); },
        overlay first={\draw[colexam,line width=.5pt] ([xshift=-0pt]title.north west) -- ([xshift=-0pt]frame.south west); },
        overlay middle={\draw[colexam,line width=.5pt] ([xshift=-0pt]frame.north west) -- ([xshift=-0pt]frame.south west); },
        overlay last={\draw[colexam,line width=.5pt] ([xshift=-0pt]frame.north west) -- ([xshift=-0pt]frame.south west); },%
    }
    \begin{testexamplebox}}
        {\end{testexamplebox}\endlist}

\AfterEndEnvironment{testexample}{\color{black}}

\DeclareMathAlphabet{\mathcal}{OMS}{cmsy}{m}{n}
\setlist[itemize]{noitemsep,nolistsep, topsep=0px, partopsep=0px, leftmargin=*}
\setlist[enumerate]{noitemsep,nolistsep, topsep=0px, partopsep=0px, leftmargin=*}
\usepackage{hyperref}
\hypersetup{
    colorlinks=true,
    linkcolor=blue,
    filecolor=magenta,      
    urlcolor=cyan
    }

\title{Vision in Action: Learning Active Perception from \\ Human Demonstrations}  

\author{
  Haoyu Xiong \textsuperscript{\rm } \hspace{0.12em}
  Xiaomeng Xu \textsuperscript{\rm } \hspace{0.12em}
  Jimmy Wu \textsuperscript{\rm } \hspace{0.12em}
  \textbf{Yifan Hou }\textsuperscript{\rm } \hspace{0.12em}
  \textbf{Jeannette Bohg} \textsuperscript{\rm } \hspace{0.12em}
  \textbf{Shuran Song} \textsuperscript{\rm } \\[1ex]
  \textsuperscript{\rm } Stanford University \\[1ex]
  \url{https://vision-in-action.github.io}
}

\begin{document}
\maketitle

\makeatletter
\renewcommand{\@makefntext}[1]{\noindent #1} % remove indentation
\makeatother
\footnotetext[0]{For any questions, please contact: haoyux.me@gmail.com}

%===============================================================================

% \begin{abstract}
% Perception is inherently active, yet most robot data collection systems rely on passive cameras, where human operators direct their gaze independently of the robot's camera viewpoint. For instance, wrist-mounted cameras are constrained by arm motions rather than driven by perceptual intent, preventing the system from capturing human perceptual behaviors such as search, exploration, and attention. This mismatch in observation between human operators and robots leads to significant learning challenges, as robots often lack task-relevant visual information. 

% We present Vision in Action (ViA), an active perception system for robot bimanual manipulation. ViA features a 6-DoF robot neck that allows the robot to mimic human-like head movements that naturally result from the combined motion of the human torso and neck. To support scalable data collection, we also introduce a teleoperation interface for novel view rendering to mitigate the motion-to-photon latency, which significantly improves operator comfort. Experiments on a range of challenging bimanual manipulation tasks demonstrate that active perception is critical for tackling real-world scenarios, where visual occlusion poses a significant challenge.

% \end{abstract}
\vspace{-0.6cm}

\begin{abstract}
    We present Vision in Action (ViA), an active perception system for bimanual robot manipulation. ViA learns task-relevant active perceptual strategies (e.g., searching, tracking, and focusing) directly from human demonstrations. On the hardware side, ViA employs a simple yet effective 6-DoF robotic neck to enable flexible, human-like head movements. 
    To capture human active perception strategies, we design a VR-based teleoperation interface that creates a shared observation space between the robot and the human operator. To mitigate VR motion sickness caused by latency in the robot's physical movements, the interface uses an intermediate 3D scene representation, enabling real-time view rendering on the operator side while asynchronously updating the scene with the robot's latest observations. 
    Together, these design elements enable the learning of robust visuomotor policies for three complex, multi-stage bimanual manipulation tasks involving visual occlusions, significantly outperforming baseline systems.
  
\end{abstract}


% \begin{figure*}[h]
%     \centering
%     \includegraphics[width=\linewidth]{fig/teaser.pdf}
%     \caption{\textbf{Active Perception} is crucial in everyday manipulation tasks. Changing viewpoint actively enables us to: 1) \textbf{reduce} visual occlusion caused by clutter (e.g., peek into the bag),
%     %
%     2) \textbf{focus} the visual attention on the critical area to infer actions (e.g., look closer when handing over a cup),
%     %
%     and 3) \textbf{increase} overall visual coverage (e.g., when searching the object in a large workspace).  
%     % https://docs.google.com/drawings/d/1vHvmSTKbgJi-LXNuNWnw-mLX-E7QgHX8wZkClhav-_4/edit?usp=sharing
%     }
%     \label{fig:teaser}
%     \vspace{-2mm}

%     %\href{https://drive.google.com/file/d/1_M_DWC6jBZRJJYNY0il3R1qiZCgYH9w-/view?usp=drive_link}{link to the keynote}, \href{https://docs.google.com/drawings/d/1TV33gLTXvV5YG2YLY9Fuec_n_ltL_2WqJwJlaUzxDGE/edit?usp=sharing}{link to the google draw}, \href{https://docs.google.com/presentation/d/1TkfQmwHv0p0XEWsxxm0EMr_aiHjXicvs/edit?usp=sharing&ouid=116441549660255333054&rtpof=true&sd=true}{link to the pptx}
    
% \end{figure*}

\section{Introduction}
\vspace{-2mm}

\begin{wrapfigure}{r}{0.5\linewidth}  
    \vspace{-6mm}
    \centering
    \includegraphics[width=\linewidth]{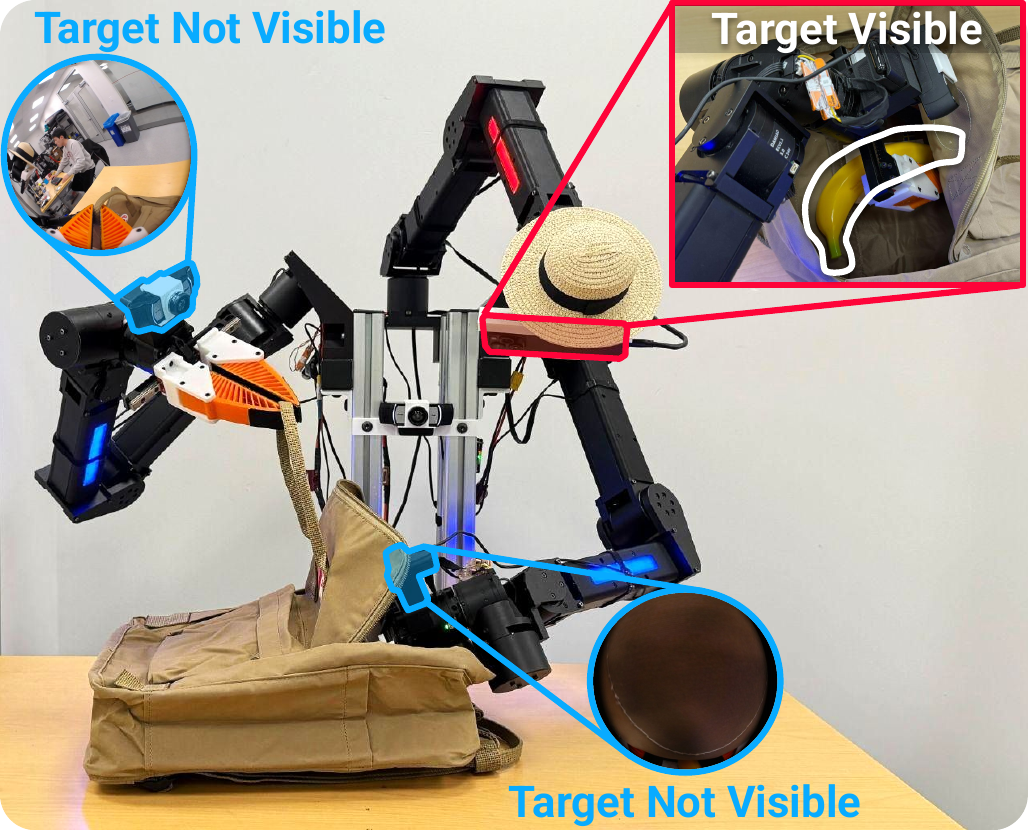}  
    \caption{\textbf{Vision in Action (ViA)} uses an \textcolor{active_camera_color}{active head camera} to search for the target object (yellow banana) inside the bag. The \textcolor{wrist_camera_color}{wrist cameras} are ineffective in this visually occluded scenario, as they are constrained by the arm motions.
    }
    \label{fig:teaser}
    \vspace{-3mm}
\end{wrapfigure}

Perception is inherently active~\cite{active-perception}. Consider the task of retrieving a banana from a bag (Fig.~\ref{fig:teaser}): one must first scan the environment to locate the bag, then peek inside to identify the banana, and finally focus on the object to determine an appropriate grasp. These deliberate viewpoint changes serve to \textbf{increase} visual coverage during the search, \textbf{reduce} occlusions caused by obstacles (\eg, the bag), and \textbf{focus} attention on action-critical regions (\eg, for grasp finding).

Yet, most robotic imitation learning systems~\cite{aloha, mobile-aloha, egomimic, dexcap, RUM, umi, thu_umi, bidex} do not incorporate active perception. 
These systems typically rely on wrist cameras~\cite{RUM, umi, thu_umi} or fixed third-person cameras~\cite{diffusion-policy}. Since wrist cameras move with the arm, their viewpoints are constrained by manipulation requirements rather than guided by perceptual objectives. 
This limitation becomes especially problematic in scenarios involving visual occlusion, where wrist cameras are often blocked by the environment and fail to capture task-relevant information necessary for accurate action inference.
Furthermore, during data collection, humans naturally shift their gaze to guide attention. However, the robot usually perceives the scene from fixed or mismatched viewpoints.
As a result, these systems fail to capture rich human perceptual behaviors such as searching, tracking, and focusing. 
This fundamental \textbf{observation mismatch}---between \textit{what the human sees} and \textit{what the robot learns from}---introduces a critical gap that ultimately hinders the learning of effective policies.
% Active perception involves devising control strategies that determine sequences of actions to minimize a loss function while maximizing information gain. 
% 
Despite its importance, active perception is often neglected in today's robotic systems due to the significant system-level challenges it introduces, including:
%
% The core reason that active perception is often avoided in today's robotic systems is that it introduces significant system-level challenges such as:

\begin{itemize}[leftmargin=3mm]
    % \item \textit{Flexible hardware requirements.} 
    \item \textit{Flexible hardware for human-like gaze control.} 
    While humans effortlessly coordinate eye, neck, and torso movements to direct their gaze in a variety of ways, replicating this capability in robots is difficult. Most robot systems today rely on fixed or constrained cameras (\eg, 2-DoF necks~\cite{Shigemi2019, 4650604, 5354572, open-television}), which limit the ability to adjust viewpoints flexibly.

     % \item \textit{Synchronized camera-eye movements.}
     \item \textit{Synchronized camera-gaze movements.}
    Virtual reality provides a powerful interface for teleoperating robots and capturing human active perception~\cite{open-television, bipasha_neckteleop}.
     However, designing an interface that synchronizes human gaze and movement of the robot camera requires precise mirroring of human motions and real-time streaming of visual feedback. Achieving this demands fast motor control and low-latency data streaming, both of which remain challenging with today's hardware.
     
    \item \textit{Scalable active perception strategies.} Human gaze is driven by top-down and bottom-up attention~\cite{human_attention,Itti_Koch01nrn,Itti_etal98pami,Tsotsos1995507}. Prior efforts to replicate human gaze behavior in robots typically relied on hand-crafted heuristics~\cite{BabakActiveVision,ArmarGaze,BOHG2012779,next_best_view}, but such strategies are difficult to generalize across diverse tasks. A more scalable approach should allow the robot to learn active perception strategies that maximize task-relevant information gain, without requiring task-specific assumptions.
\end{itemize}

% Our solution: 
In this paper, we introduce \textbf{\oursfull~(\ours)}, 
a bimanual manipulation system that learns active perception strategies directly from human demonstrations. Our system addresses the above challenges using the following design choices: 

\begin{itemize}[leftmargin=3mm]
    \item \textit{Flexible robot neck using an off-the-shelf 6-DoF arm.} Instead of replicating the intricate biomechanics of the human neck and torso through a complex design, we use an off-the-shelf 6-DoF robot arm as the robot's neck. 
    This simple yet effective approach enables human-like head movements that approximate the full range of motion produced by coordinated upper-body motion.

    \item \textit{Intermediate 3D representation to decouple human and robot motion in VR teleoperation.} 
    Instead of directly mirroring human head movements and streaming live robot camera views, we use an intermediate 3D scene representation.
    This representation enables real-time rendering of novel views based on the human's latest head pose, without requiring new observations from the robot. Consequently, the robot can be slowed down to reflect \textit{aggregated} head movements rather than every motion. This asynchronous streaming, control, and rendering bypasses the need for low-latency robot actuation and data transmission.

    \item  \textit{Shared-observation teleoperation as a scalable way to capture active perception strategies.} 
    Instead of hand-designing a gaze strategy, we let the policy learn the strategy directly from human demonstrations.
    By having the human use the same observation space as the robot---\textit{seeing what the robot sees}---we effectively capture the human's complex perceptual strategies across task stages and scenarios. 
    This enables the visuomotor policy to learn robust gaze behavior, even with straightforward behavior cloning.
\end{itemize}

To evaluate our proposed system, we perform experiments on three challenging, multi-stage bimanual manipulation tasks involving significant visual occlusions. These tasks include retrieving objects with interactive perception, rearranging cups in cluttered environments with active viewpoint switching, and precisely aligning objects using coordinated bimanual actions. 
Our experimental results highlight the critical role of active perception, with ViA outperforming baseline camera setups---such as wrist cameras and fixed chest cameras---by 45\% in success rate.
We also conducted a user study to validate the design of our teleoperation interface.
% Our experiments show that active view is useful ... 
% our formulation is good .. and scaleable ... By only changing the data, the same framework can be used for a variety of different active perception manipulation tasks,
Results are best viewed on our website: \href{https://vision-in-action.github.io}{https://vision-in-action.github.io}.

% 2. Data collection interface. if we want to design a teleoperation interface that perfectly synchronizes human demonstrator action on the robot side and streams real-time visual feedback. It requires very fast motion streaming speed that is not achievable with today's hardware. 

% 3. Inferring proper active perception strategy. Learning to control the active camera requires task-specific research -- different tasks would require different search strategies on different task stages. The classical approach often needs to design a task-specific strategy or heuristic to determine the rule of camera selection -- making it hard to scale to diverse tasks. In this work, we hope to make the system learn these strategies directly from human demonstration data. 

\vspace{-2mm}
\section{Related Work}
\vspace{-2mm}
\textbf{Active Perception and Robot Necks.}
Active perception has a long-standing history in robotics and computer vision~\cite{active-perception, BajcsyAT16, ballard1991animate, Aloimonos88, Itti_etal98pami, Itti_Koch01nrn, Tsotsos1995507}.
To investigate active perception, many artificial vision systems (\ie, humanoid necks with varying numbers of degrees of freedom) have been developed~\cite{Shigemi2019, 4650604, 5354572, toddlerbot, open-television, DarapArmRobot, ArmarHead, PAHLAVAN199241, MPISystem}. 
In this paper, we use an off-the-shelf 6-DoF arm as an active neck, with a camera mounted to the end effector. 
While the idea of using a robot arm as a neck has been explored in prior work~\cite{ArmAsNeck, ArmAsNeck2, franka_neck, bipasha_neckteleop, active_vision_rl_shaolin}, our approach integrates a novel teleoperation interface to directly control the robot neck.
%our approach builds on this concept by integrating a 3D scene interface that allows the robot to mimic human-like head movements.
%This alleviates the need for complex, custom hardware and lets us use well-understood controllers for robot arms. 
% This idea of using an arm as a neck is not new~\cite{ArmAsNeck, ArmAsNeck2,franka_neck, bipasha_neckteleop,active_vision_rl_shaolin}.
%
% In our work, the major difference is that we developed a novel 3D scene interface to control the robot neck.
The majority of prior active vision systems use various heuristics (\eg, hand-designed image filters or object detectors) to compute a measure of saliency for gaze guidance~\cite{BabakActiveVision, ArmarGaze, BOHG2012779}. There are also works that formulate the next-best-view problem in terms of uncertainty reduction~\cite{zeng_view_2020,Krainin2011, next_best_view}. 
%In these types of methods, the authors often try to estimate object shape or pose and quantify the current uncertainty about that estimate. 
In these types of methods, the objective is defined purely around a perception problem, and do not consider manipulation. 
Recent works have also investigated active vision with reinforcement learning approaches~\cite{uta_learning_to_look, active_vision_rl_katerina, active_vision_rl_2, active_vision_rl_dinesh, active_vision_rl_shaolin, spin}, though application of those methods to real-world systems remains challenging.
In contrast, our work learns bimanual manipulation and active perceptual behavior directly from real-world human demonstrations, without any task-specific assumptions.
%, resulting in search behavior and task-relevant coordination between the head camera and the robot hands such as opening a bag and turning the head to peek into it.

\textbf{Teleoperation Systems.}
Recent teleoperation works~\cite{aloha, mobile-aloha, diffusion-policy, tidybot2, robomimic, yunfan's, xu2025robopanoptes} have highlighted the potential of scaling end-to-end visuomotor policy learning.
However, existing approaches~\cite{aloha, mobile-aloha, gello, bidex} typically rely on wrist cameras~\cite{bidex, liu2023enhancing} or fixed third-person cameras~\cite{openteach_nyu}, which fail to capture the active perceptual behaviors of humans.
To overcome this human-robot observation mismatch, prior works~\cite{open-television, bipasha_neckteleop, yuke_humanoid_teleop, zoom_camera, active_vision_all_you_need, tianhao_vr_teleop, franka_neck, idp3} have explored using VR to control an active head camera~\cite{open-television}, providing immersive, first-person visual feedback through RGB video streaming.
However, these direct camera teleoperation approaches~\cite{open-television, active_vision_all_you_need} often induce motion sickness~\cite{motion_sick}, primarily due to motion-to-photon latency---the delay between the user's head movement and the corresponding visual update on the VR display~\cite{motion_to_photon}.
%accumulate due to robot control latency and data transmission latency.
% 
To address this, our method introduces an intermediate 3D scene representation that enables real-time view rendering based on the user's latest head pose, significantly reducing motion-to-photon latency.
Related to our approach, recent work~\cite{eth_nerf_teleop} introduced a VR teleoperation system that uses radiance fields to render views from a reconstructed scene. However, unlike our approach, their system lacks physical camera control. Our approach, by contrast, allows users to purposefully control the camera in VR to maintain task-relevant visibility.
%while executing closed-loop manipulation tasks.

\begin{figure*}[t]
    \centering
    \includegraphics[width=\linewidth]{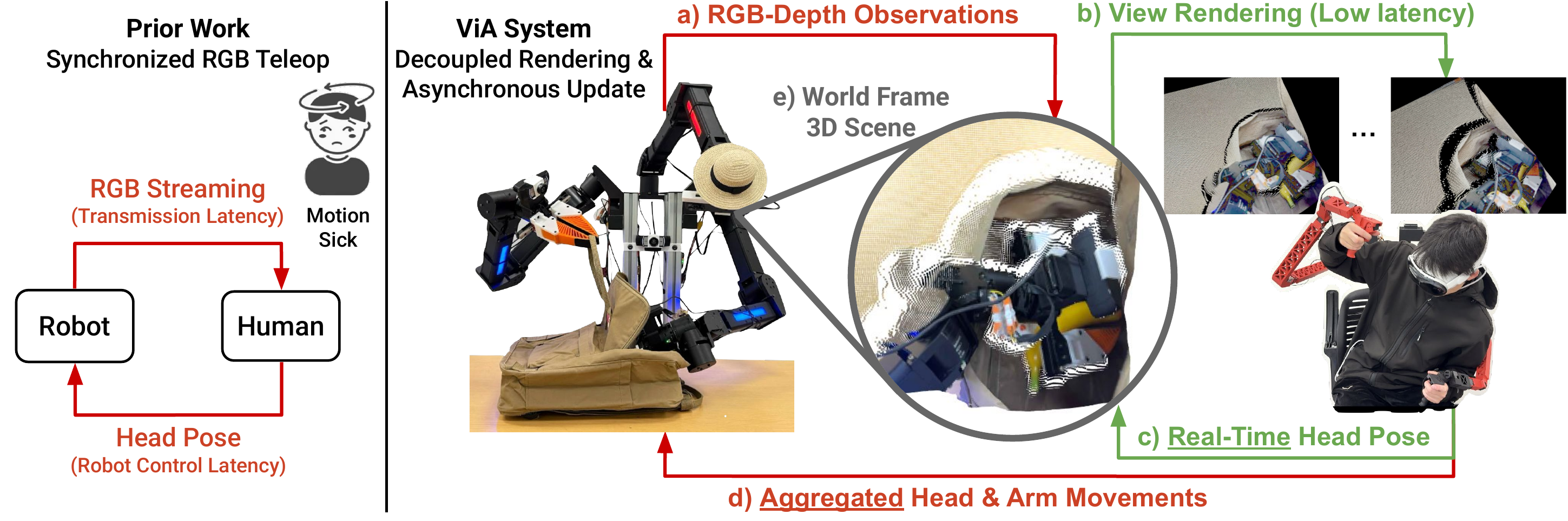}
    \vspace{-6mm}
    \caption{\textbf{VR Teleoperation Comparison.} 
    [Left] Traditional RGB streaming suffers from motion-to-photon latency due to both RGB data transmission latency and robot control latency, often leading to VR motion sickness. [Right] Our system mitigates this by: (a, e) streaming a 3D point cloud in the world frame from RGB-D data, (b, c) performing real-time view rendering based on the user's latest head pose, and (d) asynchronously updating the robot's head and arm poses. This approach enables low-latency viewpoint updates for the user.
    }
    \label{fig:teleop_system}
    \vspace{-6mm}
% {link to the pptx}    
% https://docs.google.com/drawings/d/1TV33gLTXvV5YG2YLY9Fuec_n_ltL_2WqJwJlaUzxDGE/edit?usp=sharing

\end{figure*}

\vspace{-2mm}
\section{The Vision in Action System}
\vspace{-2mm}
The ViA system features a simple yet effective robotic neck design that allows the robot to mimic human whole-upper-body movements (\S \ref{sec:hardware}).
We introduce a 3D scene interface that renders views in real-time based on the user's latest head pose. This interface asynchronously updates the underlying 3D environment while allowing the user to purposefully control the robot's active camera (\S \ref{sec:teleop}).
Finally, we propose a visuomotor policy learning framework that leverages active perception (\S \ref{sec:learning}).

\vspace{-4mm}
\subsection{Hardware Design}
\vspace{-2mm}
\label{sec:hardware}

Human active perception relies on coordinated movements of both the torso and neck to adjust head poses and acquire better viewpoints.
However, the common approach of mounting a 2-DoF neck on a static torso~\cite{open-television} provides limited flexibility and is insufficient to replicate the full range of motion.
To address this, we use an off-the-shelf 6-DoF ARX5 robot arm as a robot neck. 
This high-DoF neck design allows the robot to mimic human-like head motions that naturally result from whole-upper-body movements.
The active head camera streams real-time RGB, depth, and synchronized camera pose data. To meet these requirements, we use an iPhone 15 Pro~\cite{RUM}, mounted on the end effector of the robot neck, as the system's primary visual sensor.
To enable bimanual manipulation, we use two additional 6-DoF ARX5 robot arms~\cite{umi_on_leg} each equipped with a fin-ray parallel-jaw gripper. Each arm is mounted onto a custom 3D-printed shoulder structure.

\vspace{-4mm}
\subsection{Teleoperation Interface}
\vspace{-2mm}

\label{sec:teleop}
To collect human demonstration data, we designed a teleoperation interface that simultaneously controls both robot arms and the active neck. For the arm teleoperation, we use a full-scale bimanual exoskeleton (inspired by GELLO~\cite{gello}) that enables joint-to-joint mapping between the human user and the robot arms. For head teleoperation, we implemented a VR interface that allows the user to control the pose of the active head camera while observing visual feedback.
Our choice of VR for the head interface was motivated by the need to precisely capture human perceptual strategies. By constraining the user to use the same observations as the robot, we can record visual attention patterns that contribute to successful task execution.

\textbf{Challenge:} \textit{Exacerbated latency from physical movements.}
In the VR literature, motion-to-photon latency, also known as end-to-end latency~\cite{motion_to_photon}, refers to the delay between a user's head movement and the corresponding visual update on the display. High latency can cause discomfort or motion sickness. 
While today's consumer VR headsets achieve acceptable motion-to-photon latency (below \SI{10}{ms}) for applications like games~\cite{motion_to_photon_2}, robot teleoperation introduces an additional challenge---\textit{robot control latency}.
%
% When users move their head, we must wait for the robot to physically adjust its camera position before new visual viewpoint becomes available. This additional delay creates a significant mismatch between user's head movement and visual feedback, which can cause severe motion sickness.  
When users move their heads to teleoperate the robot's camera, there is a delay between the robot receiving and executing the command, causing the camera control to lag behind.
This additional delay creates a mismatch between the user's head movement and visual feedback, leading to potential motion sickness.  

\textbf{Solution:} \textit{View decoupling through an intermediate 3D scene representation.}  
To overcome this challenge, we decouple the user's view from the robot's view using an intermediate 3D scene representation (Fig.~\ref{fig:teleop_system}).  
This allows the user's viewpoint to update \textbf{instantly} in response to head movements (via rendering), without waiting for the robot to physically match the requested viewpoint.
While the rendered view may contain small regions with missing information (due to the delayed camera movement), \textit{the rendered view stays aligned with the user's latest head pose}---a critical factor for preserving perceptual continuity and reducing discomfort.
Concretely, the interface has three components:

%We denote $T_H$ as the robot head pose and $T_L$, $T_R$ as the robot arm poses, respectively.

\begin{itemize}[leftmargin=3mm]
    \item \textit{Point cloud construction in the world frame.} 
    We define the world frame $W$ at the fixed base of the robot neck. 
    Each RGB-D frame is transformed into this world frame using the camera intrinsics and the robot head pose (\ie, camera extrinsics \wrt the world frame) at time $t$, denoted as $^WT_\mathrm{H}(t)$. 
    This pose is computed by composing the iPhone's real-time relative pose with the initial robot head pose $^WT_\mathrm{H}(t_0)$, obtained from the robot neck's joint positions.
    The resulting point cloud $^WX(t)$ in the world frame serves as our intermediate 3D scene representation. 

    \item \textit{Low-latency view rendering.} From the point cloud $^WX(t)$, we render stereo RGB views for the VR display using the user's latest head pose in the world frame, denoted as $^WT_\mathrm{user}(t+k)$. 
    This pose is computed by transforming the VR device's head pose into the world frame $W$ with a height offset.
    This view rendering---where $k$ denotes a short time interval---enables instant visual feedback for the user. Combined with a high refresh rate (roughly 150 Hz), our system ensures smooth viewpoint updates with minimal perceived latency.
    
    \item \textit{Point cloud updating with aggregated head movements.} Finally, the robot head pose is updated to $^WT_\mathrm{H}(t+K)$ over a longer time interval $K$, using the aggregated user head pose, where $K$ is determined by the robot's control latency and is much larger than the rendering interval $k$.
    Meanwhile, the point cloud is asynchronously updated with new RGB‑D observations from the robot at a lower frequency (10\,Hz).
\end{itemize}

Overall, this teleoperation interface balances low visual latency for the user ($<$ 7\,ms) with smooth action execution on the robot side ($<$ 10\,Hz control frequency), enabling effective and practical data collection for complex manipulation tasks.

\vspace{-4mm}
\subsection{Learning Active Perception for Bimanual Manipulation}  \label{sec:learning}
\vspace{-2mm}

We design a visuomotor policy network based on Diffusion Policy~\cite{diffusion-policy} that leverages our active head camera setup to learn from human demonstrations. The policy predicts bimanual arm actions for manipulation and neck actions that mimic human active perception behaviors, conditioned on visual and proprioceptive observations. 

To enable coordinated head and arm movements, we represent the end-effector poses of the neck and arms in a common world frame. At each time step $t$, the policy receives the current RGB image observation $\mathbf{I}_t\in\mathbb{N}_0^{H \times W \times C}$ from the active head camera as the visual input, along with the proprioceptive state $\mathbf{P}_t\in\mathbb{R}^{23}$.
This state includes the end-effector poses (position and quaternion) of the neck, left arm, and right arm ($\in\mathbb{R}^7$), as well as the two gripper widths (2 scalars).

We adopt a DINOv2~\cite{oquab2023dinov2} pretrained ViT as the visual encoder for the RGB image $\mathbf{I}_t$ from the active head camera.
The 384-dimensional classification token is extracted as a compact semantic representation of the visual scene. 
The policy outputs a sequence of future actions $\mathbf{A}_t=\{a_{t+1},\ldots,a_{t+n_p}\}\in\mathbb{R}^{n_p \times 23}$, where each action consists of the future end-effector poses of the neck and arms in the world frame, as well as the gripper widths. Only the first $n_a \leq n_p$ actions are executed on the physical robot (via inverse kinematics).
We use a prediction horizon of $n_p = 16$ and an execution horizon of $n_a = 8$, with the policy operating at \SI{10}{Hz}.

%
%Following the default configuration from Diffusion Policy, we use a U-Net [cite] architecture with the DDIM scheduler for action diffusion.

\begin{figure}[t]
    \centering
    \includegraphics[width=1\linewidth]{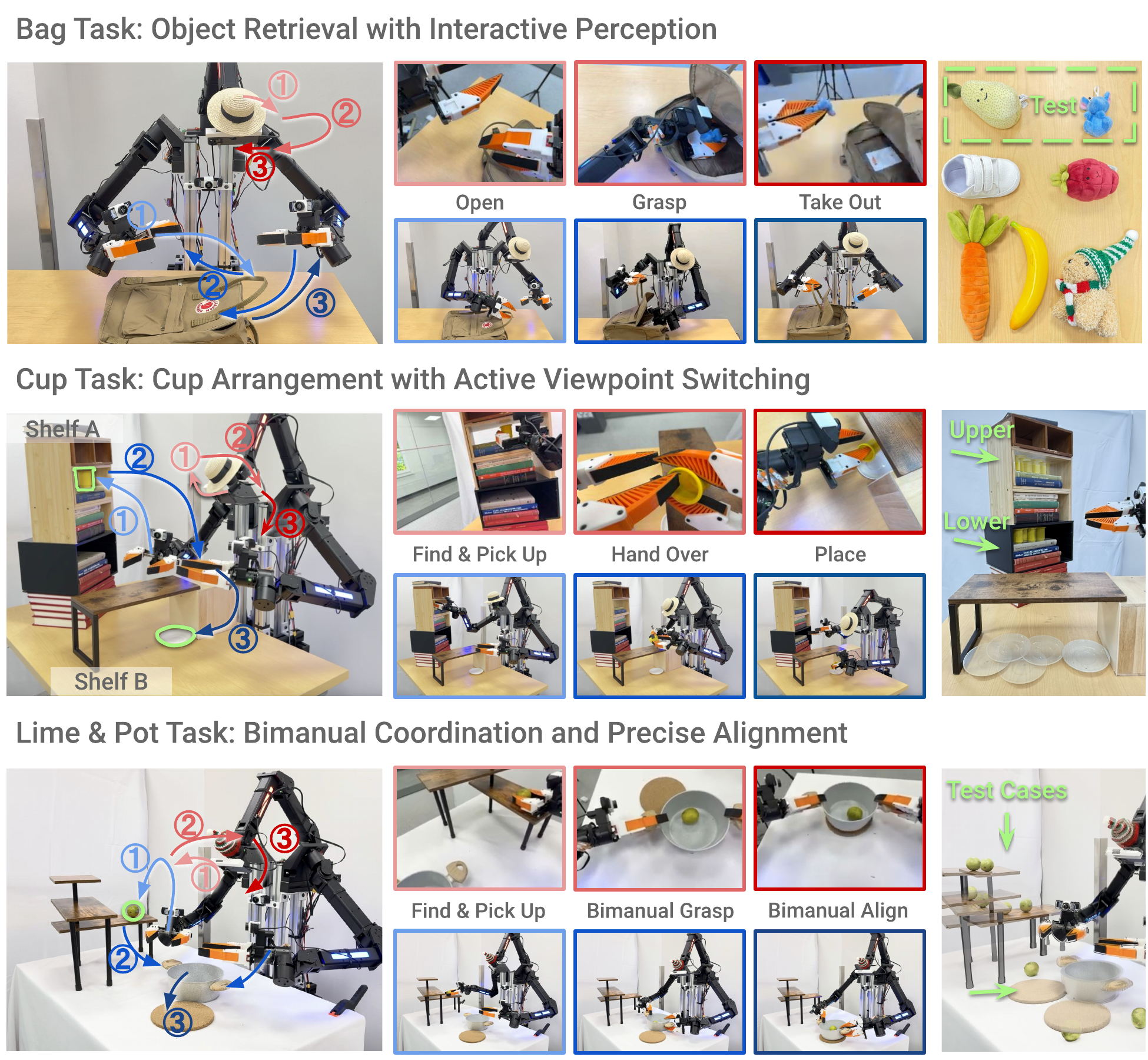}
    \vspace{-5mm}
    \caption{\textbf{Task Definitions.} We introduce three multi-stage tasks that highlight the critical role of active perception in everyday scenarios. 
    [Left] Third-person view with \textcolor{red_1}{red} \textcolor{red_2}{arrows} \textcolor{red_3}{indicating} head movements and \textcolor{blue_1}{blue} \textcolor{blue_2}{arrows} \textcolor{blue_3}{indicating} arm movements. [Middle] Active head camera views across task stages (upper row), and third-person view of robot actions (lower row). [Right] Test scenarios, including training and testing objects for the bag task, and different test configurations for the latter two tasks.
    %v1
    % https://docs.google.com/drawings/d/1ielc8_CjCaj20wxyTAFmKuhlJ-B-Mt2LycujtSgQhrE/edit?usp=sharing
    %v2
    % %https://docs.google.com/drawings/d/1MrKMj_Ukh6vVoaoWXetj7Vom6N6ALz8AiwSW9MAgopc/edit?usp=sharing 
    }
    \label{fig:task}
    \vspace{-8mm}
\end{figure}

\vspace{-2mm}
\section{Evaluation}
\vspace{-2mm}
We evaluated our system on three challenging multi-stage tasks (Fig.~\ref{fig:task}) to assess the effectiveness of various camera setups (\S\ref{sec:eval_camera}), visual representations (\S\ref{sec:eval_learning}), and teleoperation interface designs (\S\ref{sec:eval_interface}). 
For each task, we report the stage-wise success, which is defined cumulatively (\ie, success at each stage requires the successful completion of all preceding stages). Detailed task stage definitions can be found in the supplementary material.

\textit{\textbf{Bag Task}: Object retrieval with interactive perception.} The robot must (1) open a bag, (2) peek inside to locate the target object, and (3) take it out. Success requires both \textbf{\textit{active physical interaction}} (\ie, opening the bag to reduce occlusion) and \textbf{\textit{active head movement}} to inspect the bag's interior, demonstrating interactive perception. The wrist camera often suffers from limited visibility due to occlusions, whereas the active head camera can dynamically adjust its viewpoint to gather task-relevant information more effectively.
We collected 150 demonstrations with five \textit{training objects} (banana, carrot, dog, shoe, strawberry) and evaluated on two \textit{unseen test objects} (a blue elephant, a green avocado) with 5 rollouts per object---10 trials in total. For both training and evaluation, a single object is placed in the bag per trial.
\textit{\textbf{Cup Task}: Cup arrangement with active viewpoint switching.} As illustrated in Fig.~\ref{fig:task}, the robot must (1) find and pick up a cup from shelf A using its right hand, (2) hand it over to its left hand, and (3) place it on a saucer hidden beneath shelf B. Visual occlusion presents a significant challenge, requiring \textbf{\textit{active viewpoint switching}} across different stages: the cup is positioned deep within shelf A, where upper tiers obstruct wrist cameras, while the saucer is positioned beneath shelf B.
We collected 125 training demonstrations, with the cup randomly placed on either the upper or lower tier of shelf A and the saucer randomly positioned beneath shelf B. Demonstrations followed a consistent search strategy (lower tier first, then upper if needed). For evaluation, we used 10 test configurations, each run twice, resulting in 20 total rollouts.
\textit{\textbf{Lime \& Pot Task}: Bimanual coordination and precise alignment.} The robot must (1) find and place a lime into a pot, (2) lift the pot using both arms, and (3) precisely align it onto a trivet. Since the lime may appear on either side of the workspace, the robot must first \textbf{\textit{coordinate and decide}} which arm to use for grasping. Lifting the pot requires \textbf{\textit{bimanual grasping}}, and the final \textbf{\textit{precise alignment}} with the trivet is guided by the head camera to ensure precise placement.
We collected 260 demonstrations for training. For evaluation, we fixed the pot position and tested 10 different lime and trivet configurations, each tested twice for 20 total rollouts.

\begin{figure}[t]
    \centering
    \includegraphics[width=1\linewidth]{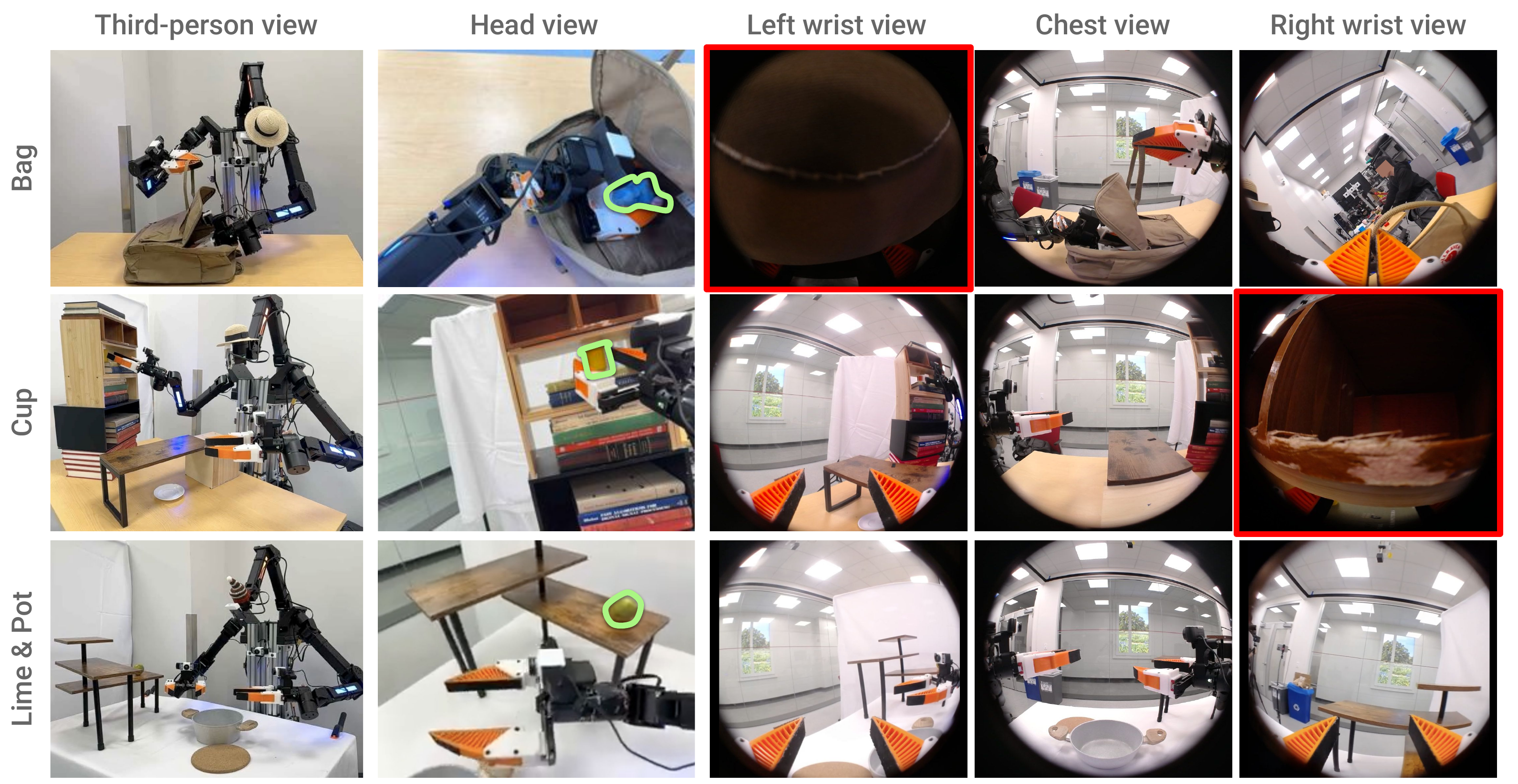}
    \vspace{-5mm}
    \caption{\textbf{Policy Learning Camera Setup Comparison.} [ViA] uses a single active head camera that dynamically adjusts its viewpoint to capture \textcolor{springgreen}{task-relevant visual information} (\eg, finding a cup hidden inside a shelf).
    In contrast, [Wrist \& Chest cameras] policy often fails due to \textcolor{brightred}{visual occlusions}. For example, in the cup task, the right wrist camera's view is blocked by the upper shelf tier, resulting in insufficient visual cues for grasping. The chest camera also fails to capture task-relevant information due to its fixed viewpoint, even when equipped with a fisheye lens.
    %v1
    % https://docs.google.com/drawings/d/1gJ5PiskrSDe2QPL4rZIWcyRP7Tq3laOfmj-z0WUpMNk/edit?usp=sharing   
    }
    \label{fig:cam_figure}
    \vspace{-4mm}
\end{figure}

\begin{figure}[t]
    \centering
    \includegraphics[width=0.99\linewidth]
    {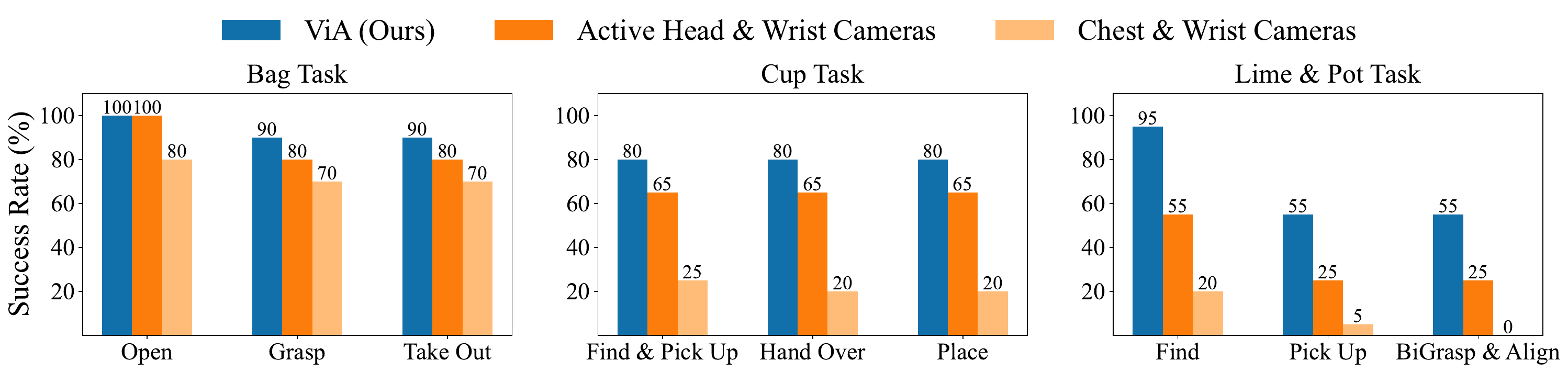}
    \caption{\textbf{Policy Learning Camera Setup Comparison Results.} We report stage-wise success rates across the three tasks to demonstrate the effectiveness of our active head camera [ViA] compared to two baseline configurations: [Active Head \& Wrist Cameras] and [Chest \& Wrist Cameras].
    }
    % change color of mix and passive
    % change other task names
    \label{fig:cam_performance}
    \vspace{-8mm}
\end{figure}

\begin{figure}[t]
    \centering
    \includegraphics[width=0.99\linewidth]{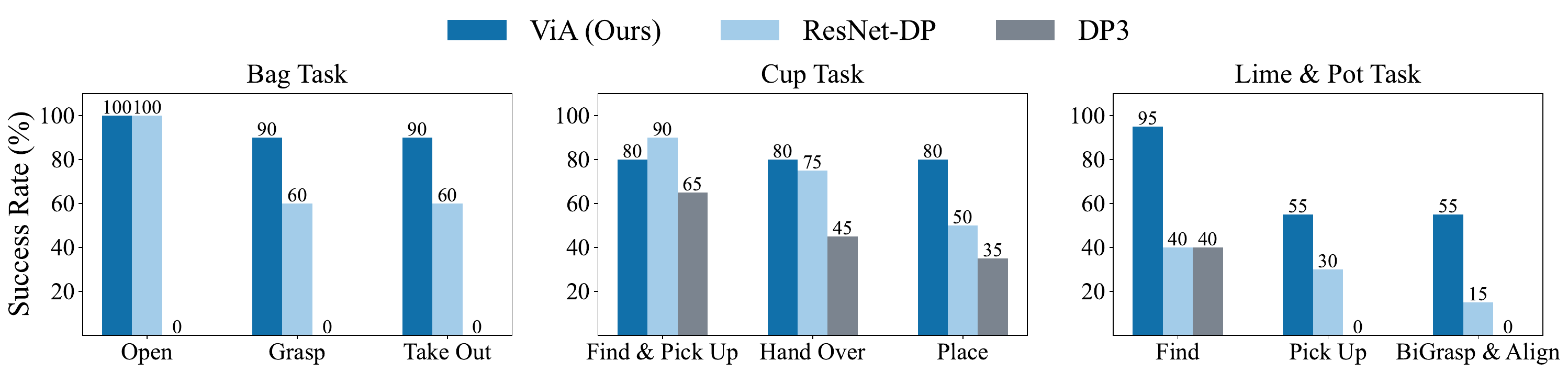}
    \vspace{-2mm}
    \caption{\textbf{Policy Learning Visual Representation Comparison Results.} We report stage-wise success rates across the three tasks to demonstrate the effectiveness of our method [ViA], in comparison to two baseline approaches: [ResNet-DP] and [DP3].}
    \vspace{-5mm}
    \label{fig:visual_rep_performance}
    \vspace{-2mm}
\end{figure}

\vspace{-4mm}
\subsection{Policy Learning Camera Setup Comparison}
\label{sec:eval_camera}
\vspace{-2mm}

\textbf{Camera Setups.} We evaluate the effectiveness of active head camera setup by comparing it with two alternative camera configurations for policy learning (Fig.~\ref{fig:cam_figure}). During data collection, all camera streams are recorded. For training, we use different combinations of these views from the same set of demonstrations, enabling a fair comparison across camera setups. 
Visual representations are extracted using a DINOv2 pretrained ViT backbone~\cite{oquab2023dinov2}.

\begin{itemize} [leftmargin=3mm]
    \item{[\textbf{ViA (Ours)}]}: Uses a single active head camera as the visual input. Details are described in \S \ref{sec:learning}. 

    \item{[\textbf{Active Head \& Wrist Cameras}]}: Combines the active head camera with two wrist cameras. Compared to [ViA], this setup includes additional wrist views as visual input. Although the teleoperator does not directly use these views, this comparison evaluates whether they provide additional useful information for policy learning.
    
    \item{[\textbf{Chest \& Wrist Cameras}]}: 
    Uses a fixed chest camera and two wrist cameras (omitting the neck). This is one of the most commonly used camera setups in current robotics systems~\cite{aloha, mobile-aloha}.

\end{itemize}

\textbf{Results.} 
As shown in Fig.~\ref{fig:cam_performance}, [ViA] consistently outperforms both alternative camera setups across all three tasks.
Surprisingly, augmenting [ViA] with additional wrist camera observations ([Active Head \& Wrist Cameras]) does not improve performance (a decrease of 18.33\% on average).
We hypothesize several reasons for this outcome: First, the active head camera alone already provides sufficient information, as the teleoperator relies solely on this view to complete the task. Thus, the visual input from the head camera alone is already task-complete. 
Second, adding wrist cameras increases input dimensionality without necessarily contributing task-relevant information.
Instead, the additional views may introduce redundant or noisy observations, especially due to frequent occlusions during manipulation.
In a low-data regime like ours, the added complexity can hinder learning by increasing the risk of overfitting or distracting the model with less informative inputs.

Next, compared to the [Chest \& Wrist Cameras] setup, it is clear that the chest and wrist cameras fail to provide sufficient task-relevant information. As shown in the second row of Fig.~\ref{fig:cam_figure}, the right wrist camera is completely occluded by the upper shelf tier during cup-grasping, while the fixed chest camera lacks visibility of the target objects altogether. In contrast, our active head camera dynamically adjusts its viewpoint, allowing the robot to gather more informative visual input and improve average task performance by 45\%.

% 2) Additional wrist-camera inputs may introduce out-of-distribution (OOD) observations. Increasing the number of input sources raises the likelihood of encountering OOD inputs, which can degrade policy performance -- especially in a low-data learning regime (like our current setting), where the model is more sensitive to distributional shifts.

\vspace{-4mm}
\subsection{Policy Learning Visual Representation Comparison}
\label{sec:eval_learning}
\vspace{-2mm}

%describe baselines 
%describe results and finding 
\textbf{Visual Representations.}
We compare [ViA] with two alternative visual representations for the policy. All policies use the same active head camera input as [ViA].
% Both the neck and bimanual arms are included in the proprioception inputs and the action outputs.
\begin{itemize}[leftmargin=3mm]
    \item{[\textbf{ViA (Ours)}]}: Uses a DINOv2~\cite{oquab2023dinov2} vision backbone for image encoding. Details can be found in \S \ref{sec:learning}.

    \item{[\textbf{ResNet-DP}]}: A baseline using a ResNet-18~\cite{resnet} backbone pretrained on ImageNet~\cite{imagenet}, integrated into diffusion policy. Input images are center-cropped to 1:1 aspect ratio and resized to $224\times224$, consistent with [ViA].
    
    \item{[\textbf{DP3}]}~\cite{ze20243d}: Uses world-frame point clouds (transformed from the active head camera) as visual input. The point cloud is cropped to the workspace and downsampled to 1,024 points. This model is trained from scratch.
\end{itemize}

\textbf{Results.}
As shown in Fig.~\ref{fig:visual_rep_performance}, our method---leveraging a pretrained DINOv2 ViT representation---achieves the highest final-stage success rate across all three tasks.
%
% explain why DP3 is so low --> e.g. bag task, DP3 can not accurately grasp the strap
% \item \textbf{Observed Behaviors \& Failure Cases.}
% \textcolor{red}{(need) to add a behavior figure}
Compared to the two baselines, [ViA] benefits from stronger semantic understanding enabled by the DINOv2 backbone. This allows the policy to actively \textit{find} the object first before initiating arm actions. For example, in the lime \& pot task, [ViA] is able to perform long-horizon active search to find the lime, before proceeding with manipulation.
%
% In contrast, a failure mode of the [DP3] baseline is hallucination, where the policy fails to interpret the scene correctly and commands the arm toward incorrect or empty regions. For example, in the cup task, [DP3] often commands the arm to go an empty section, where the target cup is placed in another section.
%
In contrast, a common failure mode of the [DP3] baseline is hallucination, where the policy misinterprets the scene and issues incorrect actions. For example, in the cup task, [DP3] often directs the arm to an empty section of the shelf, failing to identify the actual cup location.
[DP3] also completely fails on the bag task due to the imprecise grasping of the bag handle in the open stage.
We hypothesize that this is due to the limited semantic capacity of the [DP3] representation, which is trained from scratch and lacks pretrained visual priors. 

% We hypothesize this is due to the lack of semantic understanding in raw point cloud inputs.
% (We include a visualization of this failure in Fig.~\ref{fig:hallucination_example}.) 

\vspace{-4mm}
\subsection{Teleoperation Interface Comparison}
\label{sec:eval_interface}
\vspace{-2mm}

\begin{wrapfigure}{r}{0.5\textwidth}
    \vspace{-7mm}
    \centering
    \includegraphics[width=0.5\textwidth]{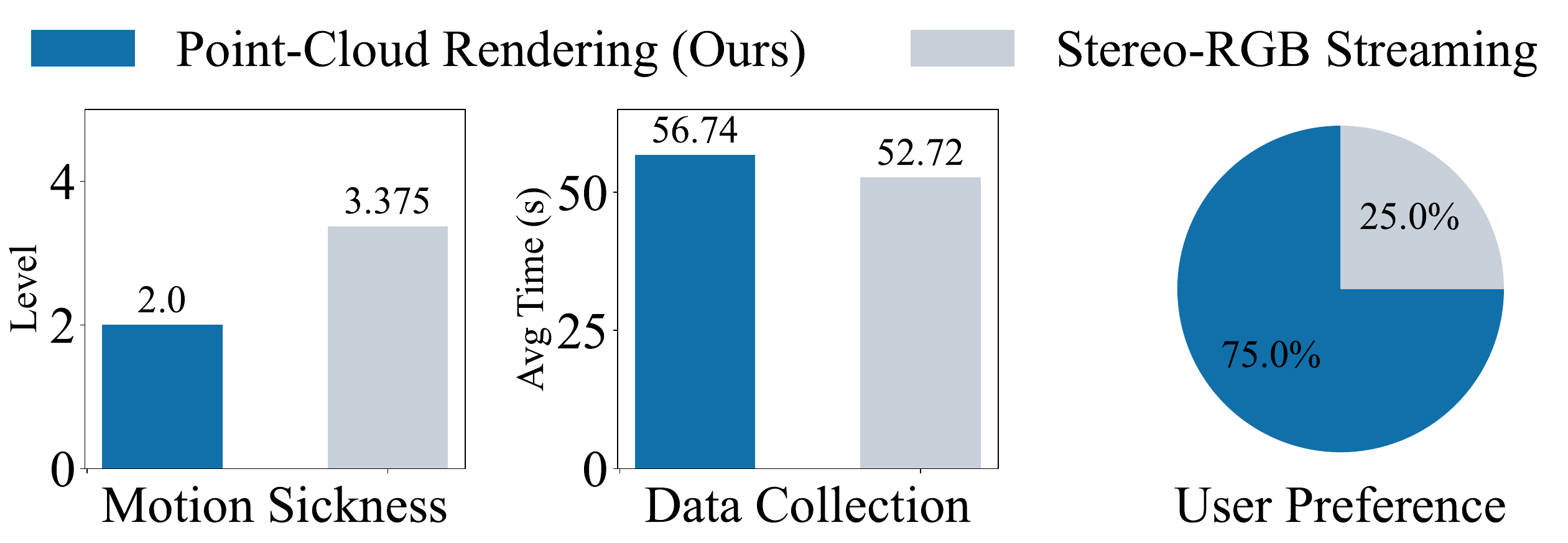}
    \vspace{-5mm}
    \caption{\textbf{Teleoperation Interface Comparison.} We evaluate our teleoperation interface design based on three metrics: reported levels of motion sickness, average duration to complete each demonstration, and overall user preference.} 
    \label{fig:teleop}
\end{wrapfigure}

In this experiment, we evaluate our VR teleoperation interface by comparing our system---which uses a point cloud rendering method---with a conventional RGB streaming approach~\cite{open-television, bipasha_neckteleop, active_vision_all_you_need}.
We conducted a user study with 8 participants of varying levels of experience with VR and robot teleoperation. All participants were first-time users of both systems and were unaware of which system corresponded to our proposed design.
For each participant, the order of system usage was randomized and labeled as System A and System B.
Participants were asked to perform the cup task using both systems. Each session included a 5-minute practice period followed by a data collection phase in which participants provided 3 demonstrations.
We recorded the completion time for each demonstration and gathered user feedback through a post-session experience survey.
\textbf{Results.}
As shown in Fig.~\ref{fig:teleop}, while our point cloud rendering method results in slightly longer data collection times compared to stereo RGB streaming, it significantly reduces motion sickness. As a result, 6 out of 8 participants reported a preference for our system.

\vspace{-3.5mm}
\section{Conclusion} 
\vspace{-3.5mm}
\label{sec:conclusion}
%\todo{}
% We developed Vision in Action (ViA), an active perception system for bimanual robot manipulation. 
%
The ViA system features a simple yet effective neck design that allows the robot to mimic human-like head movements.
We developed a teleoperation interface that renders real-time views based on the user's latest head pose, while asynchronously updating the scene by controlling the robot's active head camera to gather task-relevant information.
%
%A user study validated that our interface significantly reduces motion sickness compared to baselines.
%
% We conducted a careful user study to validate that our 3D scene interface significantly reduces motion sickness compared to baseline methods. 
%
For evaluation, we introduced three challenging multi-stage tasks involving significant visual occlusion for policy learning.
Experimental results highlight the importance of active perception, with ViA significantly outperforming baseline setups.

\newpage 
\section{Limitations} 
\label{sec:limitations}

This work explores how active perception can be learned from human demonstrations, taking an initial step toward that broader goal. While our system shows promising results, it also presents several limitations. Below, we outline three areas for future improvement and research.

\textbf{Teleoperation Interface Design.}
Our 3D scene interface enables real-time view rendering from a point cloud, transformed from single-frame RGB-D data. However, due to noisy depth sensing and incomplete scene reconstruction, the resulting visualization can be lower in fidelity compared to traditional RGB video streaming. As a result, users often require practice to adapt and may find it challenging to perform fine-grained manipulation tasks.
One possible future direction is to explore dynamic scene fusion and rendering techniques~\cite{4dgs}, which remain an important yet challenging problem.
In addition, wearable devices such as AR glasses~\cite{egomimic} hold great promise for capturing human active perception in everyday tasks, potentially removing the need for physical robot teleoperation during data collection.

\textbf{Hardware Design.}
Using an off-the-shelf 6-DoF arm as a neck is a simple yet effective solution. However, this design may fall short in replicating the full complexity of human whole-body movements.
We see exciting opportunities in optimizing hardware designs that enable more human-like behaviors and facilitate learning from humans~\cite{xu2024dynamics, moma_codesign, toddlerbot}.
We are also interested in upgrading our current tabletop systems to mobile manipulation platforms, where active perception becomes more challenging and better reflects real-world scenarios.

\textbf{Policy Learning Design.}
There are several opportunities to improve the design choices in our current policy learning framework.
First, we believe it is valuable to explore representation learning that fuses observations from all cameras into a shared space, rather than simply concatenating encoded features, which may improve overall task performance in the future.
Second, our policy is not yet conditioned on language. Incorporating reasoning presents a promising direction, especially when combined with active perception. Natural language instructions often imply high-level goals, spatial cues, object relationships, or temporal dependencies. By conditioning policies on language, robots can better interpret human intent, dynamically adjust perception strategies, and disambiguate between similar visual scenes. 
Finally, tasks involving search—such as our Lime \& Pot task—require memory. The robot must search for the lime before executing any arm actions, and ideally remember which areas have already been searched to avoid repetitive behavior. Our current policy learning framework does not support such memory capabilities.
% %
% Currently, we use the absolute end-effector pose with respect to the world frame as the action space. 
% %
% However, this action space design limits spatial generalization. For instance, if all training data involves a fixed table height, the policy may struggle to generalize to different table heights in novel environments. This limitation becomes even more significant in mobile manipulation scenarios, where the robot body is also in motion—highlighting the importance of designing a more generalizable action space.
% %

% A figure for Comparison on camera: 
% 1. Task 1: 
% grasp part comparison 
% Placing: other camera cannot see plot 
% 2. Task 2: 
% Alignment for the pot, other camera cannot see 
% Active search 

% 
% chest camera can also see the alignment, so we probly won't make this point in the paper

% 3. Task 3: Pick stage 

% Other:

% Task pot: left and right hand pick
% task pot: long horizon active searching
%===============================================================================

% \clearpage
% The acknowledgments are automatically included only in the final and preprint versions of the paper.
\acknowledgments{
This work was supported in part by the Toyota Research Institute, NSF Award \#2143601, \#2037101, and \#2132519, the Sloan Foundation, Stanford Human-Centered AI Institute, and Intrinsic. 
The views and conclusions contained herein are those of the authors and should not be interpreted as necessarily representing the official policies, either expressed or implied, of the sponsors. 

We would like to thank  ARX for the ARX robot hardware. 
We thank Yihuai Gao at Stanford for his help on the ARX robot arm controller SDK.
We thank the help from Ge Yang at MIT and Xuxin Cheng at UCSD for their help and discussion of VR.
%
% ReaLab helpful discussions.
We thank Max Du, Haochen Shi, Austin Patel, Zeyi Liu, Huy Ha, Mengda Xu, 
% SVL helpful discussions.
Yunfan Jiang, 
%
% Karen Lab discussions
Ken Wang, Yanjie Ze for their helpful discussions.
We thank all the volunteers who participated in and supported our user study.

}

%===============================================================================

\newpage
\bibliography{example}  % .bib

\end{document}